\documentclass[conference]{IEEEtran}
\IEEEoverridecommandlockouts
\usepackage{cite}
\usepackage{amsmath,amssymb,amsfonts}
\usepackage{algorithmic}
\usepackage{graphicx}
\usepackage{textcomp}
\usepackage{xcolor}
\usepackage{url}
\usepackage{comment}
\usepackage{hyperref}
\usepackage{orcidlink}
\def\BibTeX{{\rm B\kern-.05em{\sc i\kern-.025em b}\kern-.08em
    T\kern-.1667em\lower.7ex\hbox{E}\kern-.125emX}}
\begin{document}

\title{An Example for Domain Adaptation Using CycleGAN
}

\author{\IEEEauthorblockN{Yanhua Zhao
}}

\maketitle

\begin{abstract}
Cycle-Consistent Adversarial Network (CycleGAN) is very promising in domain adaptation. In this report, an example in medical domain will be explained. We present struecture of a CycleGAN model for unpaired image-to-image translation from microscopy to pseudo H\&E stained histopathology images. 
\end{abstract}


\section{Introduction}
Light sheet microscopy enables the rapid volumetric imaging of biological tissues using fluorescence labelling. Although this technique preserves fine structural detail, histopathological interpretation is traditionally performed on haematoxylin and eosin (H\&E)-stained images. Overcoming this gap between the two modalities can significantly improve interpretability for pathologists.

Converting fluorescence microscopy images to H\&E-like appearance would enable pathologists to visualize fluorescence data in an interpretable format, facilitate integration with existing analysis tools, and support comparative analysis between modalities. Traditional image translation methods require paired training data, which is often impractical due to technical constraints and sample preparation differences in histopathology.

This report presents a Cycle-Consistent Adversarial Network (CycleGAN) approach for unpaired image-to-image translation from fluorescence microscopy to pseudo H\&E stained images. The architecture employs ResNet-based generators with residual blocks and PatchGAN discriminators, trained using adversarial, cycle-consistency, and identity losses to ensure realistic translation while preserving morphological structures.

The contributions of this work include: 

\begin{itemize}
    \item a novel application of using CycleGAN for domain adaptation in histopathology,
    \item a framework operating without paired training data,
    \item preservation of morphological structures while adopting H\&E-like color characteristics. 
\end{itemize}

\section{Method Selection for Virtual Staining}

Several approaches exist for virtual staining. Classical methods and supervised models are limited by their inability to handle complex textures or by the need for paired datasets. More recent techniques include Contrastive Unpaired Translation (CUT), which in many cases outperforms CycleGAN by reducing hallucinations, preserving tissue structure better, and enabling faster training, though it remains relatively complex and requires a large number of images. Advanced conditional diffusion models can produce very realistic textures and high visual quality, but they are data intensive, difficult to control in terms of morphology, and generally more complex than needed for this task. CycleGAN remains the most common choice for domain adaptation because it does not require paired data and learns a mapping between two image domains, making it well suited for translating fluorescence images into realistic H\&E images using unpaired training with datasets such as MHIST. 

CycleGAN [1] was selected because it enables bidirectional domain translation without paired supervision and preserves structural consistency through cycle consistency loss.

\section{Data Preparation and Preprocessing}

Source domain is microscopy images. To create RGB inputs, we combine the channels using a semantic color mapping: channel 1 maps to the blue channel, channel 2 maps to the green channel, and the red channel receives a 30\% contribution from channel 2. Before combination, each channel is normalized to reduce the impact of outliers and intensity variations, then scaled to [0, 1] and converted to RGB format.

For Target Domain, the MHIST dataset \cite{b3} is used (available at \url{https://bmirds.github.io/MHIST/}
), which contains high quality H\&E stained histopathology images suitable for learning realistic color and texture distributions. The dataset contains 3,152 H\&E images.

All images are resized to 256×256 using bilinear interpolation. During training, data augmentation includes random horizontal and vertical flips (p=0.5) and color jitter (brightness and contrast ±0.1). Images are converted to tensors and normalized to [-1, 1] using mean=0.5 and std=0.5 to match the generator’s tanh output range.

\begin{figure*}[htbp]
    \centering
    \includegraphics[width=0.9\linewidth]{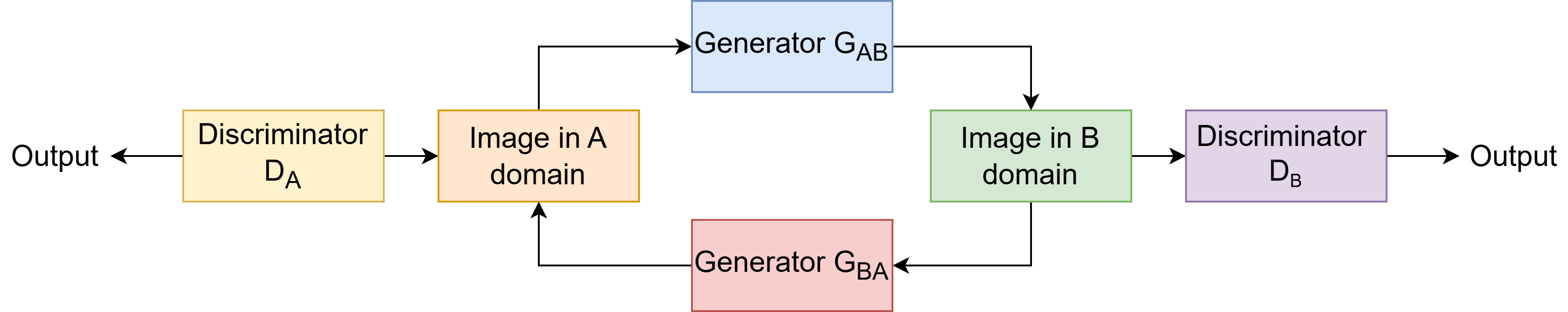}
    \caption{Overview of the proposed virtual H\&E generation pipeline. A is Source Domain, B is Target Domain.}
    \label{fig:system}
\end{figure*}

\section{Model Architecture}
The model includes two generators and two discriminators trained adversarially, as shown in Figure \ref{fig:system}.

\subsection{Generator Architecture}
Each generator is a ResNet based encoder-decoder. The encoder: initial 7×7 convolution with reflection padding (64 filters), two downsampling blocks (stride 2) increasing filters to 256, and 9 residual blocks with instance normalization and ReLU. The decoder: two transposed convolution upsampling blocks (stride 2) reducing filters back to 64, and a final 7×7 convolution with tanh activation producing RGB outputs in [-1, 1]. Reflection padding is used throughout to reduce boundary artifacts.
\subsection{Discriminator Architecture}
Each discriminator has: an initial 4×4 convolution (stride 2, 64 filters) with LeakyReLU (0.2), two additional downsampling blocks (stride 2) increasing filters to 256 with instance normalization, a final 4×4 convolution (stride 1) producing a single-channel output map, and LeakyReLU (0.2) throughout. This design focuses on local patch-level realism.
\subsection{Cycle-Consistent Architecture}
The model includes two generators: G\_A2B (fluorescence → H\&E) and G\_B2A (H\&E → fluorescence), and two discriminators: D\_A (fluorescence domain) and D\_B (H\&E domain). The forward cycle (A→B→A) and backward cycle (B→A→B) enforce cycle consistency, ensuring that translating an image to the target domain and back reconstructs the original image. Identity mappings are computed for both domains to preserve domain specific characteristics when the input is already in the target domain.

\section{Results}

\begin{figure}[htbp]
    \centering
    \includegraphics[width=0.45\textwidth]{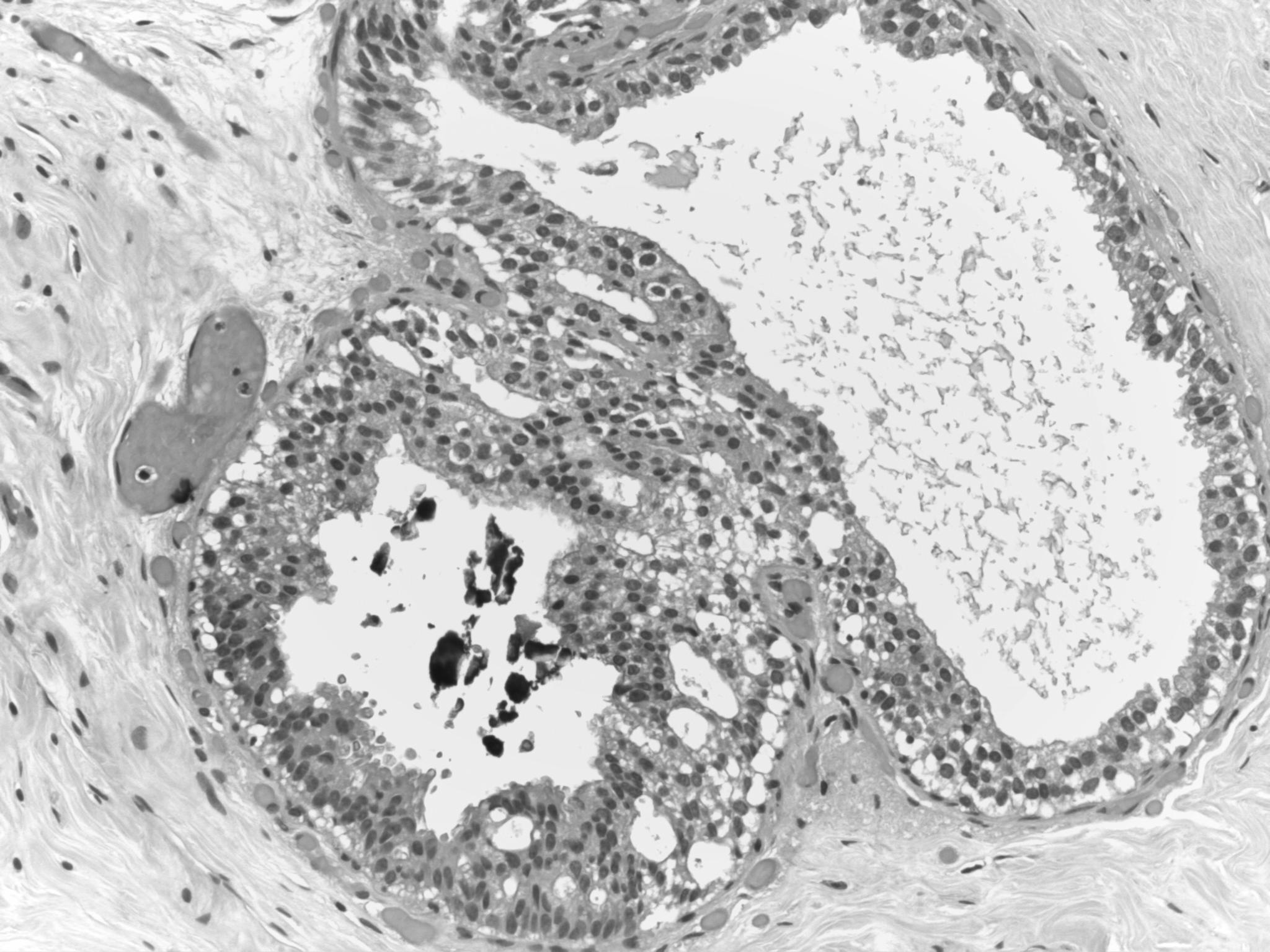}
    \hfill
    \\
    \vspace{12pt}
    \includegraphics[width=0.45\textwidth]{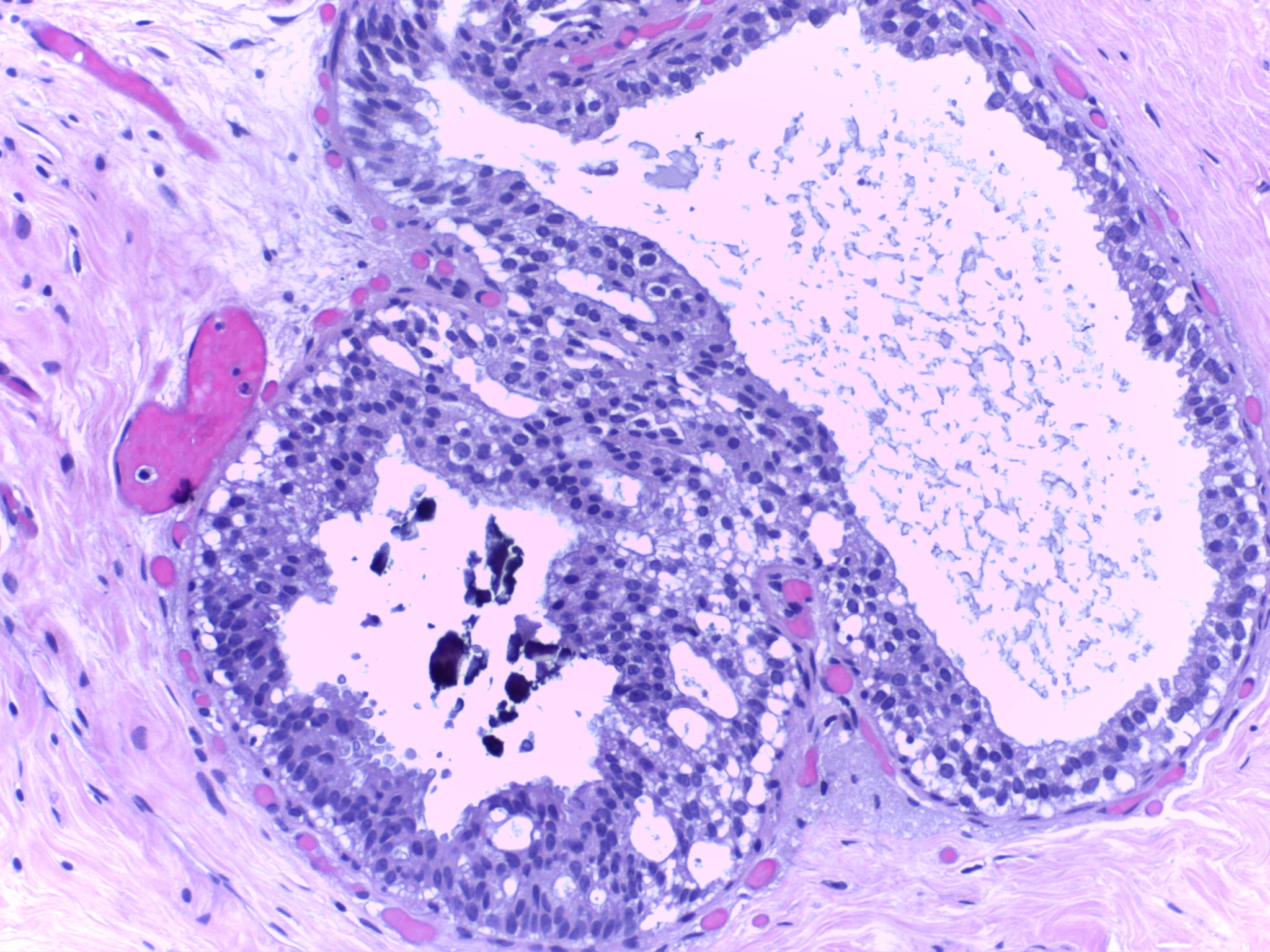}
    \caption{Top: Example of input. Bottom: Schematic example of output \cite{b4}. }
    \label{fig:grayscale_png}
\end{figure}

Figures \ref{fig:grayscale_png} shows representative example outputs from CycleGAN, included solely to demonstrate domain adaptation feasibility between fluorescence microscopy and the H\&E domain.

\section{Limitations and Possible Improvements}

A primary limitation of this work is the relatively small size of source domain dataset. This limited diversity may restrict the model’s ability to generalize across different tissue regions and imaging conditions, increasing the risk of overfitting. Although the H\&E target domain contains 3,152 images, the quality and diversity of these reference images strongly influence the realism of the generated virtual stains. Biases or limited variability in the H\&E dataset can lead to suboptimal color mapping and texture representation.

Another limitation is the lack of explicit confidence estimates or quality control mechanisms for the generated virtual H\&E images. The model produces outputs deterministically without providing uncertainty measures, making it difficult to automatically identify low-quality results. As a consequence, some generated images may exhibit artifacts, unrealistic textures, or color inconsistencies, particularly during early training stages or when processing challenging tissue structures.

Several improvements can be explored in future work. Incorporating quantitative evaluation metrics, such as structural similarity measures and task-specific metrics tailored to virtual H\&E image quality, would enable more objective assessment of model performance. Expanding the training dataset with additional fluorescence samples and more diverse H\&E images would likely improve robustness and generalization. Finally, alternative unpaired translation models, such as Contrastive Unpaired Translation (CUT), could be investigated to reduce hallucination effects and further improve structural preservation in virtual staining results.

\vspace{12pt}


\begin{thebibliography}{00}

\bibitem{b2} J. Vasiljević, Z. Nisar, F. Feuerhake, C. Wemmert, and T. Lampert,
``CycleGAN for Virtual Stain Transfer: Is Seeing Really Believing?'',
\emph{Artificial Intelligence in Medicine}, vol. 133, p. 102420, 2022.


\bibitem{b3}Jerry Wei, Arief Suriawinata, Bing Ren, Xiaoying Liu, Mikhail Lisovsky, Louis Vaickus, Charles Brown, Michael Baker, Naofumi Tomita, Lorenzo Torresani, Jason Wei, Saeed Hassanpour, “A Petri Dish for Histopathology Image Analysis”, International Conference on Artificial Intelligence in Medicine (AIME), 12721:11-24, 2021.

\bibitem{b1} Dubey, Shikha, et al. "Structural cycle gan for virtual immunohistochemistry staining of gland markers in the colon." International Workshop on Machine Learning in Medical Imaging. Cham: Springer Nature Switzerland, 2023.

\bibitem{b4} Aresta, G., Araújo, T., Kwok, S., Chennamsetty, S. S., Safwan, M., Alex, V., … \& Aguiar, P. (2019). Bach: Grand challenge on breast cancer histology images. Medical image analysis, 56, 122-139.

\end{thebibliography}
\end{document}